\documentclass{article}
\usepackage{ijcai24}

\usepackage{times}
\usepackage{soul}
\usepackage{url}
\usepackage[hidelinks]{hyperref}
\usepackage[utf8]{inputenc}
\usepackage[small]{caption}
\usepackage{graphicx}
\usepackage{amsmath}
\usepackage{amsthm}
\usepackage{booktabs}
\usepackage{algorithm}
\usepackage{algorithmic}
\usepackage[switch]{lineno}

\usepackage{graphicx}
\usepackage{amsmath}
\usepackage{amssymb}
\usepackage{booktabs}
\usepackage{enumitem}
\usepackage{algorithm}
\usepackage{algorithmic}
\usepackage{array}
\usepackage{makecell}
\usepackage{graphicx}
\usepackage{multirow}
\usepackage{booktabs}
\usepackage{rotating}
\usepackage{color}
\usepackage{subcaption}

\title{Multiple Population Alternate Evolution Neural Architecture Search}

\author{
    Juan Zou\textsuperscript{1}, Han Chu\textsuperscript{1}\thanks{Corresponding author}, Yizhang Xia\textsuperscript{1}, Junwen Xu\textsuperscript{1}, Yuan Liu\textsuperscript{1}, Zhanglu Hou\textsuperscript{1}
    \affiliations
    \textsuperscript{1}Hunan Engineering Research Center of Intelligent System Optimization and Security, Xiangtan University
    \emails
    chuhan@smail.xtu.edu.cn
}

\begin{document}

\maketitle

\begin{abstract}
    The effectiveness of Evolutionary Neural Architecture Search (ENAS) is influenced by the design of the search space. Nevertheless, common methods including the global search space, scalable search space and hierarchical search space have certain limitations. Specifically, the global search space requires a significant amount of computational resources and time, the scalable search space sacrifices the diversity of network structures and the hierarchical search space increases the search cost in exchange for network diversity. To address above limitation, we propose a novel paradigm of searching neural network architectures and design the Multiple Population Alternate Evolution Neural Architecture Search (MPAE), which can achieve module diversity with a smaller search cost. MPAE converts the search space into $L$ interconnected units and sequentially searches the units, then the above search of the entire network be cycled several times to reduce the impact of previous units on subsequent units. To accelerate the population evolution process, we also propose the the population migration mechanism establishes an excellent migration archive and transfers the excellent knowledge and experience in the migration archive to new populations. The proposed method requires only 0.3 GPU days to search a neural network on the CIFAR dataset and achieves the state-of-the-art results.
\end{abstract}

\section{Introduction}

Neural Architecture Search (NAS) is a method that utilizes machine learning techniques to automatically search and optimize neural network architectures. The goal of NAS is to improve the efficiency or performance of deep learning models by searching a smaller or better neural network. NAS typically employs various techniques, including reinforcement learning~\cite{zoph2016neural}, gradient descent~\cite{liu2018darts} and evolutionary computation~\cite{xie2017genetic}, to discover optimal neural network architectures that yield superior performance. This method has been widely used in image classification~\cite{sun2019evolving}, object detection~\cite{liang2021opanas} and natural language processing~\cite{klyuchnikov2022bench}.

In evolution-based NAS, the performance of the model depends on the design of the search space. Nevertheless, common methods including the global search space, scalable search space and hierarchical search space have certain limitations. A widely adopted concept is the global search space that directly search the entire structure of the neural network to identify the most optimal configuration~\cite{zhang2022evolutionary}. However, this approach demands significant computational resources and time due to the expansive nature of the search space.
The second strategy involves dividing the network into fundamental cells and constructing a more intricate network by stacking these cells together. The typical approach is cell-based NAS~\cite{zoph2018learning}, which first explores cells and then combines the cells by stacking and splicing the searched cells to create a complete network. The cell-based NAS reduces the search complexity and enhances the adaptability of the structure. However,~\cite{tan2019mnasnet} point out that the cell-based stacking structure compromises the diversity of network structures and does not fully take into account the characteristics and limitations of various parts of the entire network.
Thirdly, ~\cite{tan2019mnasnet} proposed a hierarchical search space with a fixed macro structure, it allows different layers to utilize different resolution blocks within the network structure. However, this approach incurs additional costs when searching for different layer structures.
In order to balance the diversity of network structures with search costs, we introduce an alternate search approach. In a network formed by stacking $N$ cells, We transform the search space into a space of $L$ interconnected cells, and search the $N$ cell spaces sequentially. This approach revisits the neural architecture search paradigm from a split perspective. The alternate search method splits a global search space into multiple small global search spaces to divide and conquer, greatly simplifying the complexity of the problem and improving search efficiency.

To address the challenges posed by a larger search space and the issue of lengthy network encoding~\cite{zhang2022evolutionary} in the application of evolutionary algorithms to NAS, We propose a multi-population alternate evolutionary (MPAE) framework based on alternate search.
Unlike other methods~\cite{tian2020multipopulation} that utilize multiple populations to explore various equivalent Pareto-optimal solutions, the MPAE employs multiple populations to partition the extensive search space into subsets based on the network structure. Each population then samples from a different search space subset. This approach not only ensures an equal distribution of the lengthy network encoding across each population but also simplifies the search space that each individual population needs to explore.

To leverage the knowledge accumulated during the optimization process of each network module, we investigate the impact of inter-module migration within the network. Drawing inspiration from the repetitive structures commonly found in CNNs (e.g. the repetitive modules in Inception~\cite{szegedy2017inception}, ResNet~\cite{he2016deep} and NASNet~\cite{zoph2018learning}), we observe that there exists a similarity in the excellent network structures of adjacent layers. Building upon this insight, we introduce the population migration mechanism, which harnesses the retained knowledge and experience of each population to expedite the evolutionary process.
The population migration process includes three aspects: maintaining the migration archive, determining the number of migrating individuals for each population and selecting the migrating individuals. These migration archives are not solely selected based on accuracy, but also include potential excellent architectures based on the Pareto frontier. Based on the phenomenon that structural differences in network modules increase with distance, the populations choose different numbers of migrating individuals from the migration archive set of other populations based on their proximity. To ensure population diversity, the selection of migrating individuals is determined by the similarity values between individuals and the migrating population.

In summary, our contributions are as follows:
\begin{itemize}
\item {\bf Multiple Populations Alternate Evolution Framework:} We proposed Multiple Populations Alternate Evolution (MPAE) framework to simplify the search space and meet the diversification needs of modules with a small search cost. MPAE uses multiple populations to split a global search space into multiple small global search spaces to divide and conquer, which greatly reduces the complexity of a single population to search.
\item {\bf Migration Mechanism:} we explore the effects caused by the migration of each module of the network to each other, and propose the population migration mechanism. The population migration mechanism constructs a migration archive set, and uses the Euclidean distance and similarity between network modules as indicators to select migrating individuals. This mechanism not only ensures the diversity of the population, but also makes full use of the knowledge retained in the optimization process of each network module to assist the search.
\end{itemize}

\section{Related Work}
The influence of the NAS search space on search outcomes~\cite{yu2019evaluating,zhang2020you} has been found to be substantial. As NAS algorithms have evolved, the design of the search space has also evolved. Initially, in the early stages of NAS development, the common practice~\cite{xie2017genetic,sun2019evolving,real2017large} was to directly search the entire structure of the neural network. However, this global search approach proved to be resource-intensive and time-consuming. To improve search efficiency, scalable network architectures~\cite{zoph2018learning,liu2018darts,pham2018efficient,yang2020cars} were introduced. This approach involves constructing networks by repetitively stacking identical components, allowing for the extension of network components that were searched on smaller datasets to larger ones, but the scalable network architecture has sacrificed consequently the diversity of the network. And the studies~\cite{girish2019unsupervised,zeiler2014visualizing} have highlighted that convolutional networks learn distinct features at different layers. This realization emphasized the importance of block diversity in influencing the performance of network models. Consequently, a hierarchical search space~\cite{cai2018proxylessnas,tan2019mnasnet,wu2019fbnet} with a fixed macrostructure was adopted, it enabling different layers to utilize different resolution blocks within the network structure.
However, these algorithms incurred an additional cost when searching for different layer structures. In contrast, our alternation evolution paradigm not only benefits from the scalable network structure but also allows for the search of different layer structures without incurring additional costs.

\section{Method}

\subsection{Overview}
\begin{figure*}[ht]
  \centering
  \includegraphics[width=0.85\textwidth]{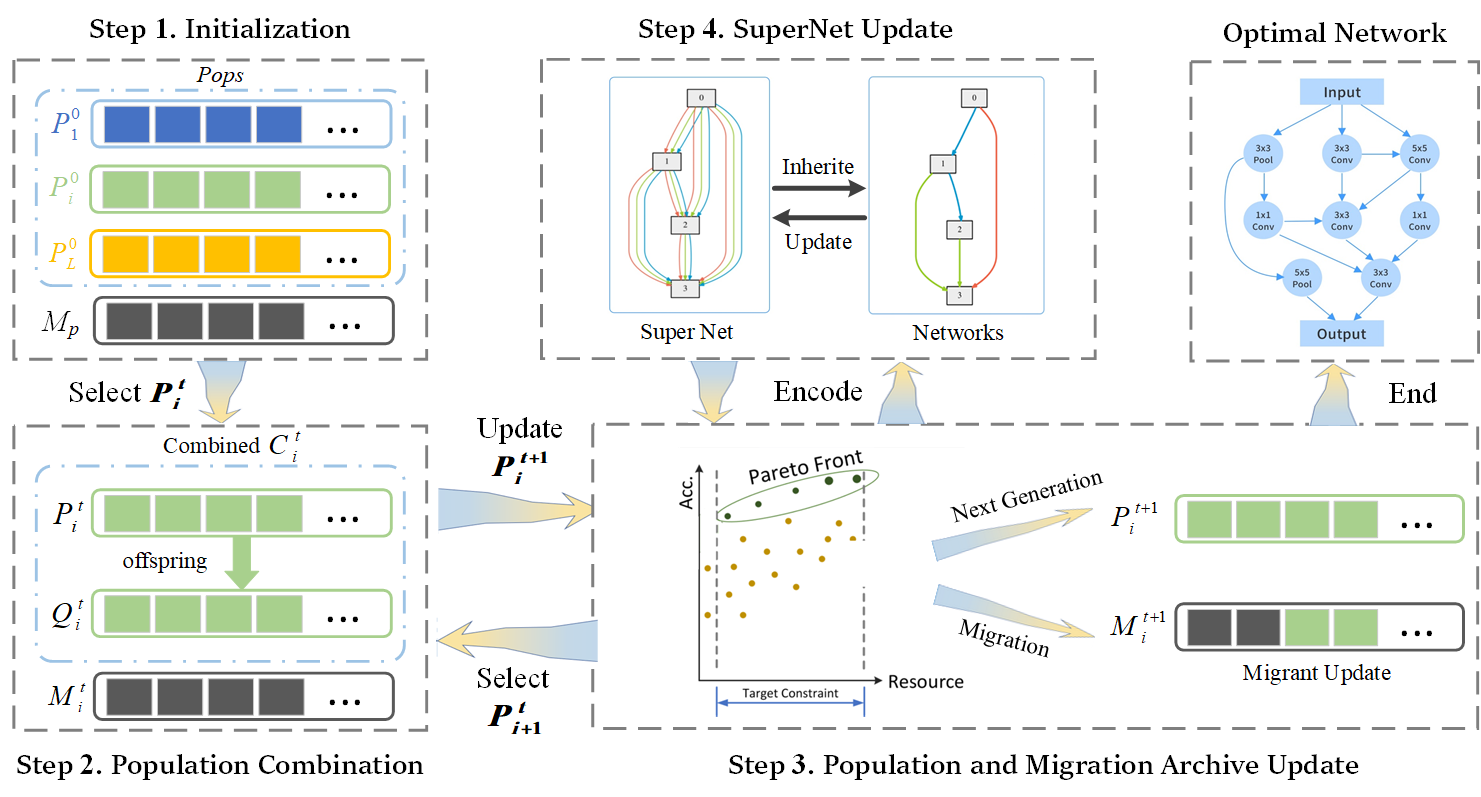}
  \caption{{\bf Overall process of MPAE.}
  Step 1 of the algorithm is to initialize multiple populations and immigrant archives. Step 2 is to generate offspring populations and merge them with parent populations and migrating populations. Step 3 involves select the parent populations for the next generation through multi-objective environmental selection and updating the immigrant archives. Step 4 is when the populations inherit the weights of the super network during evolution and update the weights of the super network after training for a few epochs. Finally, the optimal network is obtained through the search process.
  }
  \label{fig:Overview}
\end{figure*}
We develop a multi-population alternate evolution strategy to search for neural architectures, $i.e.$ MPAE.
The search phase of MPAE consists of four parts, as depicted in Figure 1.
The pipeline of the MPAE scheme is as follows:

\paragraph{Initialization:}
Initially, according to the $L$-layer structure of the neural network, the search space $\mathbf{\mathit{}}\mathcal{A}$ is divided into subsets $\mathbf{\mathit{}}\mathcal{A}_{l}$. Then $N$ candidate sub-networks are randomly selected from the search space $\mathbf{\mathit{}}\mathcal{A}_{l}$ to form the population $pop_l$. In practice, the subset $\mathbf{\mathit{}}\mathcal{A}_{l}$ of the search space is much smaller than the whole set $\mathbf{\mathit{}}\mathcal{A}$, and we believe that such limited-sized subsets of the search space are naturally easier for search algorithms to handle to avoid the dilemma of large search spaces.

\paragraph{Population Conbination:} After selecting the current evolutionary population $P_{i}^{t}$, the population $P_{i}^{t}$ generates the offspring population $Q_{i}^{t}$ through genetic operations. Then the two populations and the migration population $M_{i}^{t}$ generated through the population migration mechanism (section 3.4) are merged into the combined population $C_{i}^{t}$.

\paragraph{Population and Migration Archive Update:} All individuals in the combined population $C_{i}^{t}$ are decoded into corresponding neural structures, inherit the supernet weights and undergo training evaluation. Then multi-objective environmental selection is used to select outstanding individuals on the Pareto front from the combined population $C_{i}^{t}$, and generate the next parent population $P_{i}^{t+1}$ and the migration archive set $M_{i}^{t+1}$.

\paragraph{SuperNet Update:} The supernet parameter is warmed up for $m$ epochs before starting the search. In the search phase, supernet training and population evolution are carried out alternately, and supernet will sample candidate individuals from the population to participate in training and update supernet parameters.



\subsection{Initialization}
\begin{figure*}[ht]
  \centering
  \includegraphics[width=0.85\textwidth]{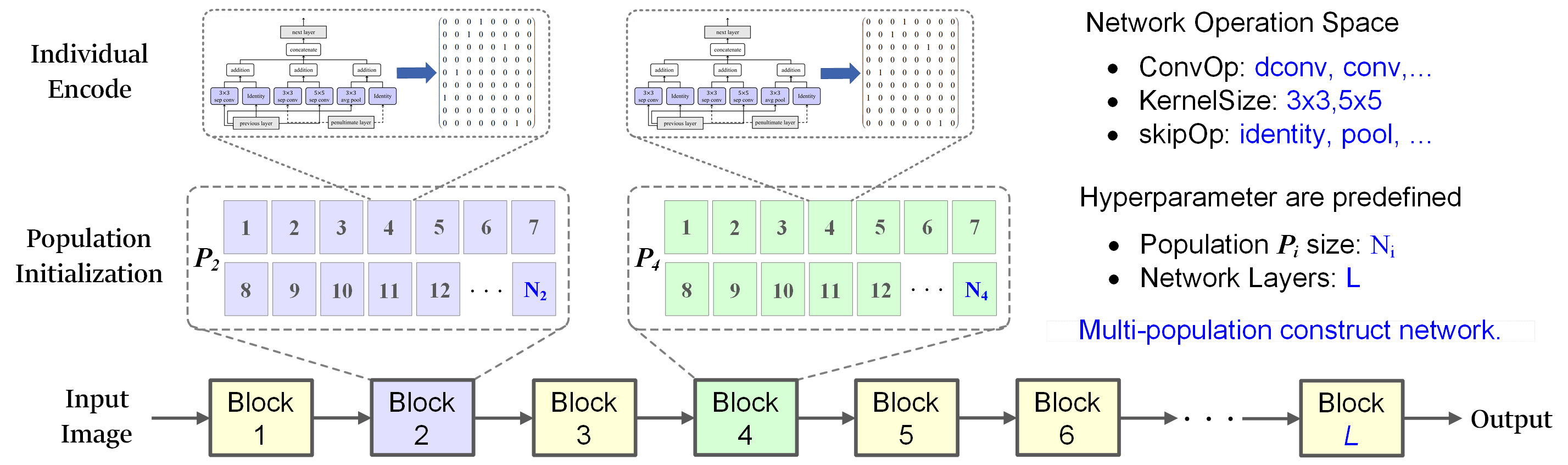}
  \caption{{\bf Multi-Populations Initialization.} Individual encode refers to the transformation of networks sampled from the search space into a two-dimensional matrix through encode strategy. The initialization of population $P_i$ is the process of generating $N_i$ network individuals. The initialization of multiple populations is the process of creating $L$ populations based on the number of layers $L$ in the complete network. It is important to note that the $L$ individuals sampled from multiple populations together constitute a complete network.}
  \label{fig:Initialization}
\end{figure*}
The Multi-Populations Initialization is shown in Figure 2. Since neural architectures typically use feedforward structures, we represent the entire search space pool $\mathbf{\mathit{}}\mathcal{A}$ containing $L$ layer blocks as a directed acyclic graph (DAG). Assuming that the number of possible operation combinations in each block is $N$, the total number of operations contained in the entire search space pool is $N^{L}$. In order to simplify the complexity of the problem, we divide the entire search space pool $\mathbf{\mathit{}}\mathcal{A}$ into $L$ subsets $\mathbf{\mathit{}}\mathcal{A}_{l}$ and use $L$ populations to represent these sub-search space pools, ${\textstyle \sum_{l=1}^{L}} {\mathbf{\mathit{}}\mathcal{A}_{l}}$. The total number of operations included in the simplified $L$-layer neural network is $N\times L$.

The initialization of multiple populations is the process of creating $L$ populations respectively based on the number of layers $L$ of the complete network. Each population $P_i$ randomly generates $N_i$ individuals and the generation of individuals corresponds to the network structure being encoded into a binary genome. The blocks corresponding to each layer in the search space we use are the same as the Cell in DARTS~\cite{liu2018darts}. DARTS uses a cell-level search space, where the cell uses four nodes to represent the sum of feature maps and edges to represent operations. The difference from DARTS is that the network structure parameters we define are $\{0,1\}$ binary encode.

\subsection{Supernet Training}
\paragraph{Subnetwork Sampling:}
Given $L$ populations $P_{1:L}$, sampling a complete subnetwork $s_i$ from a supernet $\mathcal{S}$ is achieved by sample $L$ individuals from the multipopulation $P_{1:L}$. Therefore, the sampling process of the subnetwork $s_i$ can be defined as follows:
\begin{equation}
  s_{i}=\sum_{l\in\mathcal{L}} decode \left(p_{i}^{l}\right) 
\end{equation}
Where $L$ represents the layer number index set of the $L$-layer sub-network, and also represents $L$ populations. Where $p_{i}^{l}$ denotes the individual $p_i$ sampled from the $l_{th}$ population.

\paragraph{Weight Update:}
The weight $W_{\mathcal{S}}$ of the supernet $\mathcal{S}$ is updated by training on the sampling subnetwork $s_i$. It is worth noting that the initial weight $W_{\mathcal{S}}(s_i)$ of subnetwork $s_i$ is obtained by inheriting the weight $W_{\mathcal{S}}$ of supernet $\mathcal{S}$ through weight sharing. The optimization of weights $W_{\mathcal{S}}$ can be expressed as follows:
\begin{equation}
  W_{\mathcal{S}}=\underset{W_{\mathcal{S}}}{\arg \min } \sum_{s_i\in\mathcal{S}} \mathcal{L}_{train}\left(\mathcal{M}\left(s_i, W_{\mathcal{S}}(s_i)\right)\right)
\end{equation}
where $\mathcal{L}_{train}(\cdot)$ represents the cross-entropy loss function on the training set, and $\mathcal{M}\left(s_i, W_{\mathcal{S}}(s_i)\right)$ denotes the model with subnetwork structure $s_i$ and weights $W_{\mathcal{S}}(s_i)$.
In the subnetwork sampling stage, we sample every possible architecture equally and jointly optimize the loss functions of these architectures, so each individual is sampled with $p_i \sim Bernoulli(0.5)$.

\begin{table*}[ht]\small
  \centering
  \scalebox{1}{
    \renewcommand\arraystretch{0.8}
    \tabcolsep=0.3cm
    \begin{tabular}{cc|ccc|ccc|cc}
    \toprule
    \multicolumn{2}{c|}{\multirow{2}[4]{*}{Algorithm}} & \multicolumn{3}{c|}{CIFAR 10} & \multicolumn{3}{c|}{CIFAR 100} & \multicolumn{2}{c}{\multirow{2}[4]{*}{Search space}} \\
\cmidrule{3-8}    \multicolumn{2}{c|}{} & ACC(\%) & \# P(M) & GDs   & ACC(\%) & \# P(M) & GDs   & \multicolumn{2}{c}{} \\
    \midrule
    \multicolumn{2}{c|}{CNN-GA~\cite{sun2020automatically}} & 96.78 & 2.9   & 35    & 79.47 & 4.1   & 40    & \multicolumn{2}{c}{Global} \\
    \multicolumn{2}{c|}{SI-EvoNet~\cite{9268174}} & 97.31 & 1.84  & 0.458 & 79.16 & 0.99  & 0.813 & \multicolumn{2}{c}{Global} \\
    \multicolumn{2}{c|}{AE-CNN~\cite{sun2019completely}} & 95.3  & 2     & 27    & 77.6  & 5.4   & 36    & \multicolumn{2}{c}{Global} \\
    \multicolumn{2}{c|}{AE-CNN+E2EPP~\cite{sun2019surrogate}} & 94.7  & 4.3   & 7     & 77.98 & 20.9  & 10    & \multicolumn{2}{c}{Global} \\
    \multicolumn{2}{c|}{EPCNAS-C~\cite{9930866}} & 96.93 & 1.2   & 1.1   & 81.67 & 1.29   & 1.1   & \multicolumn{2}{c}{Global} \\
    \midrule
    \multicolumn{2}{c|}{AmoebaNet-A~\cite{real2019regularized}} & 96.66±0.06 & 3.1   & 3150  & 81.07 & 3.1   & 3150  & \multicolumn{2}{c}{Scalable} \\
    \multicolumn{2}{c|}{NASNet-A~\cite{zoph2018learning}} & 97.35 & 3.2   & 1800  & 82.19 & 3.2   & 1800  & \multicolumn{2}{c}{Scalable} \\
    \multicolumn{2}{c|}{BlockQNN-S~\cite{zhong2020blockqnn}} & 96.7  & 6.1   & 90    & 82.95 & 6.1   & 90    & \multicolumn{2}{c}{Scalable} \\
    \multicolumn{2}{c|}{PNAS~\cite{liu2018progressive}} & 96.59±0.09 & 3.2   & 225   & 82.37 & 3.2   & 225   & \multicolumn{2}{c}{Scalable} \\
    \multicolumn{2}{c|}{MdeNAS~\cite{zheng2019multinomial}} & 97.45 & 3.8   & 0.16  & 82.39 & 3.8   & 0.16  & \multicolumn{2}{c}{Scalable} \\
    \multicolumn{2}{c|}{RENAS~\cite{chen2019renas}} & 97.12±0.02 & 3.5   & 6     & -     & -     & -     & \multicolumn{2}{c}{Scalable} \\
    \multicolumn{2}{c|}{GDAS-NSAS~\cite{9247292}} & 97.25±0.08 & 3.5   & 0.4   &  81.98±0.05 & 3.5   & 0.4   & \multicolumn{2}{c}{Scalable} \\
    \multicolumn{2}{c|}{ENAS~\cite{pham2018efficient}} & 97.11 & 4.6   & 0.5   & 81.09 & 4.6   & 0.5   & \multicolumn{2}{c}{Scalable} \\
    \multicolumn{2}{c|}{WPL~\cite{pmlr-v97-benyahia19a}} & 96.19 & -     & -     & -     & -     & -     & \multicolumn{2}{c}{Scalable} \\
    \multicolumn{2}{c|}{Random (F=64)~\cite{bender2018understanding}} & 95.6  & 6.7   & -     & -     & -     & -     & \multicolumn{2}{c}{Scalable} \\
    \multicolumn{2}{c|}{SNAS~\cite{xie2018snas}} & 97.15±0.02 & 2.8   & 1.5   & 79.91 & 2.8   & 1.5   & \multicolumn{2}{c}{Scalable} \\
    \multicolumn{2}{c|}{PC-DARTS~\cite{xu2021partially}} & 97.43±0.07 & 3.6   & 0.1   & 82.89 & 3.6   & 0.1   & \multicolumn{2}{c}{Scalable} \\
    \multicolumn{2}{c|}{DARTS~\cite{liu2018darts}} & 97.00±0.14 & 3.3   & 1     & 82.46 & 3.3   & 1     & \multicolumn{2}{c}{Scalable} \\
    \multicolumn{2}{c|}{CARS~\cite{yang2020cars}} & 97.38 & 3.6   & 0.4   & 82.72 & 3.6   & 0.4   & \multicolumn{2}{c}{Scalable} \\
    \multicolumn{2}{c|}{PDARTS~\cite{chen2019progressive}} & 97.5  & 3.4   & 0.3   & \textcolor[RGB]{70,70,70}{84.08} & 3.6   & 0.3   & \multicolumn{2}{c}{Scalable} \\
    \multicolumn{2}{c|}{NSGANetV1-A3~\cite{lu2020multiobjective}} & \textcolor[RGB]{70,70,70}{97.78} & 2.2   & 27    & 82.77 & 2.2   & 27    & \multicolumn{2}{c}{Scalable} \\
    \midrule
    \multicolumn{2}{c|}{TS-ENAS~\cite{zou2023ts}} & 97.57 & 2.98   & 0.9     & \textcolor[RGB]{70,70,70}{82.69}     & 3.35     & 0.9     & \multicolumn{2}{c}{Hierarchical} \\
    \multicolumn{2}{c|}{Proxyless NAS~\cite{cai2018proxylessnas}} & \textbf{97.92} & 5.7   & 1500     & -     & -     & -     & \multicolumn{2}{c}{Hierarchical} \\
    \midrule
    \multicolumn{2}{c|}{MPAE-A} & 97.35 & 2.8   & 0.3   & 82.74 & 2.9   & 0.3   & \multicolumn{2}{c}{Our} \\
    \multicolumn{2}{c|}{MPAE-B} & 97.39 & 3.2   & 0.3   & 83.45 & 3.3   & 0.3   & \multicolumn{2}{c}{Our} \\
    \multicolumn{2}{c|}{MPAE-C} & 97.51 & 3.7   & 0.3   & \textbf{84.12} & 3.6   & 0.3   & \multicolumn{2}{c}{Our} \\
    \bottomrule
    \end{tabular}}
  \caption{Comparison of MPAE with other NAS and OSNAS methods in terms of testing classification accuracy ($ACC$), number of parameters ($P$), and search cost (in terms of GPU Days, $GDs$) on CIFAR10 and CIFAR100 benchmark datasets. Symbol “–” indicates that no results have been reported on the dataset in the original paper.}
  \label{CIFAR}%
\end{table*}%

\subsection{Multi-Populations Alternative Evolutionary}
In the evolutionary search phase, instead of using individuals in a single population to encode the global neural network, we simplify the encoding of the global network structure by multiple populations as a way to reduce the complexity of the search problem. The algorithm uses populations to represent different modules of the global network and alternatively optimizes each module. The details of the algorithm are described in Algorithm 1.
\begin{algorithm}[ht]
	\renewcommand{\algorithmicrequire}{\textbf{Input:}}
	\renewcommand{\algorithmicensure}{\textbf{Output:}}
	\caption{Multi-Populations Alternative Evolutionary Search Algorithm}
	\label{alg1}
	\begin{algorithmic}[1]
            \REQUIRE SuperNet $\mathcal{S}$, network layers size $L$, population size $N$, iterations size $E_{evo}$, archives size $m$, multi-objectives $\{\mathcal{F}_{1},...,\mathcal{F}_{M}\}$.
		\STATE $P_{1:L} \gets$ Initialize multiple populations with the size $N$.
		\STATE $archives_{1:L} \gets$ Initialize immigration archive via select $m$ individuals from $P_{1:L}$.
            \STATE Warm up the SuperNet $\mathcal{S}$ for $E_{warm}$ epochs.
            \FOR {$t = 1,..., E_{evo}$}
                \FOR {$l = 1,..., L$}
                \STATE Training the individuals in $P_{t}^{l}$ by using the {sampled training strategy} on $D_{train}$
                \STATE $Q_{t}^{l}\gets$ Generate sub-population via genetic manipulation
                \STATE $P_{t+1}^{\ l}\gets$ Select $N$ individuals from $P_{t}^{l}\cup Q_{t}^{l}\cup M_{t}^{l}$ through multi-objective environment selection
                \STATE $\{\mathcal{F}_{1},...,\mathcal{F}_{M}\} \gets$ Evaluate the multi-objectives of individuals in $P_{t}^{l}\cup Q_{t}^{l}\cup M_{t}^{l}$ 
                \STATE $archive_{l}\gets$ Select $x$ individuals from $P_{t+1}^{\ l}$ through multi-objective environment selection
                \ENDFOR 
            \ENDFOR 
		\ENSURE Architectures $S=\left\{s_{1}^{\left(E_{\text{evo}}\right)},\ldots,s_{P}^{\left(E_{\text{evo}}\right)}\right\}$
	\end{algorithmic}  
\end{algorithm}

Algorithm 1 presents the framework of our proposed MPAE approach. Initially, multiple populations are initialized using the proposed genetic encoding strategy (line 1), and immigration archives are initialized for each population. Evolutionary processes commence after a warm-up period of the supernet for a few epochs and continue until a predefined termination criterion, such as reaching the maximum number of generations. Finally, individuals on the Pareto front are selected and decoded into their corresponding networks for final training.

During the evolution process, all populations undergo an alternate evolution in cycles. A multi-objective evaluation is first performed on all individuals in the population based on weight sharing. Subsequently, new offspring are generated through our designed genetic manipulation. The immigrant population is then generated using the proposed proximity selection operation. Finally representatives are selected from the existing individuals, newly generated offspring, and the immigrant population to form the next generation population and participate in subsequent evolution.

\subsection{Migration Mechanism}
Early research~\cite{girish2019unsupervised,zeiler2014visualizing} shows that different layers of the network learn different features, the optimal structure of different layers is not necessarily the same.
However, we consider that the structure of excellent networks in adjacent layers is similar, and the repeated structures commonly used in CNNs (such as repeated modules in Inception~\cite{szegedy2017inception}, ResNet~\cite{he2016deep} and NASNet~\cite{zoph2018learning}) also show this phenomenon. Therefore, we add a population migration mechanism to the algorithm, introduce excellent and potential individuals from other populations during the evolution of the current population, and make full use of the knowledge gained from the evolution of other populations to promote the evolution of the current population.

The migration mechanism determines the number of individuals migrated according to the adjacent distance of each population. The adjacent distance of the population is the difference between the corresponding network layer serial numbers of each population. At the same time, the migrating individuals of a population are selected based on the degree of similarity between the individuals and the population. The degree of similarity between individual $Gen_a$ in population $P_a$ and population $P_b$ is expressed by the following equation:
\begin{equation}
  \operatorname{sim}\left(\operatorname{Gen}_{a}, \operatorname{P}_{b}\right)=\frac{\operatorname{Gen}_{a} \times \sum_{i=1}^{D} \operatorname{Gen}_{b}^{i}}{D\cdot Len\left(Gen\right)}
\end{equation}
where $\sum_{i=1}^{D}\operatorname{Gen}_{b}^{i}$ represents the first $D$ best individual gene codes (the sum of each target value is the smallest) in the population $P_b$.
The smaller the value of $\operatorname{sim}\left(\operatorname{Gen}_{a}, \operatorname{P}_{b}\right)$, the less similar the migration individual $Gen_a$ selected in population $P_a$ is to population $P_b$, with the aim of increasing the diversity of population $P_b$ while ensuring individual fitness.

\section{Experiment}
In this section, we present the experimental results of our proposed algorithm compared to other competitors on benchmark classification datasets, CIFAR and ImageNet. The results include accuracy ($\%$), parameter count (M), and search time (GDs). GDs refers to the number of days required to search for the specified neural network on a single GPU. Additionally, we categorize the other competitors into three categories based on the search space: Scalable, Global, and Hierarchical. The Scalable category primarily constructs networks by stacking the best-performing blocks obtained from the search. The Global category directly searches for complete networks on the dataset without the need for repeated block proxies. The Hierarchical category can be regarded as a special type of Global category, which searches each network block separately, allowing network blocks with different structures to construct the complete network.
\paragraph{Evolution Details.} To demonstrate the effectiveness of our method, we adopt the same cell-level search space as DARTS~\cite{liu2018darts}. MPAE builds an $L$ layer supernet to assist evolution and adopts alternate search to explore $L$ different cell structures of the supernet. We set the supernet layers and populations size  to 20 and the individuals size of population to 64. During the evolutionary search, the crossover and mutation rate were both set to 0.25. During the genetic manipulation, each node has a 0.25 probability of undergoing connection crossover and has a 0.25 probability of being randomly reassigned.

\begin{table*}[ht]\small
  \centering
    \renewcommand\arraystretch{0.8}
    \tabcolsep=0.2cm
    \begin{tabular}{cccccccccccccc}
    \toprule
    \multicolumn{2}{c}{\multirow{2}[4]{*}{Algorithm}} & \multicolumn{4}{c}{ImageNet test accuracy (\%)} & \multicolumn{2}{c}{\multirow{2}[4]{*}{\# Params(M)}} & \multicolumn{2}{c}{\multirow{2}[4]{*}{\makecell[c]{Search Cost\\(GPU-days)}}} & \multicolumn{2}{c}{\multirow{2}[4]{*}{\makecell[c]{Search\\dataset}}} & \multicolumn{2}{c}{\multirow{2}[4]{*}{Search space}} \\
\cmidrule{3-6}    \multicolumn{2}{c}{} & \multicolumn{2}{c}{Top-1} & \multicolumn{2}{c}{Top-5} & \multicolumn{2}{c}{} & \multicolumn{2}{c}{} & \multicolumn{2}{c}{} & \multicolumn{2}{c}{} \\
    \midrule
    \multicolumn{2}{c}{SI-EvoNAS~\cite{9268174}} & \multicolumn{2}{c}{75.8} & \multicolumn{2}{c}{92.59} & \multicolumn{2}{c}{4.7} & \multicolumn{2}{c}{0.458} & \multicolumn{2}{c}{CIFAR10} & \multicolumn{2}{c}{Global} \\
    \multicolumn{2}{c}{AmoebaNet~\cite{real2019regularized}} & \multicolumn{2}{c}{75.7} & \multicolumn{2}{c}{92.4} & \multicolumn{2}{c}{6.4} & \multicolumn{2}{c}{3150} & \multicolumn{2}{c}{CIFAR10} & \multicolumn{2}{c}{Scalable} \\
    \multicolumn{2}{c}{DARTS~\cite{liu2018darts}} & \multicolumn{2}{c}{73.3} & \multicolumn{2}{c}{91.3} & \multicolumn{2}{c}{4.7} & \multicolumn{2}{c}{4} & \multicolumn{2}{c}{CIFAR10} & \multicolumn{2}{c}{Scalable} \\
    \multicolumn{2}{c}{NASNet-A~\cite{zoph2018learning}} & \multicolumn{2}{c}{74} & \multicolumn{2}{c}{91.6} & \multicolumn{2}{c}{5.3} & \multicolumn{2}{c}{1800} & \multicolumn{2}{c}{CIFAR10} & \multicolumn{2}{c}{Scalable} \\
    \multicolumn{2}{c}{PNAS~\cite{liu2018progressive}} & \multicolumn{2}{c}{74.2} & \multicolumn{2}{c}{91.9} & \multicolumn{2}{c}{5.1} & \multicolumn{2}{c}{225} & \multicolumn{2}{c}{CIFAR10} & \multicolumn{2}{c}{Scalable} \\
    \multicolumn{2}{c}{MdeNAS~\cite{zheng2019multinomial}} & \multicolumn{2}{c}{73.2} & \multicolumn{2}{c}{–} & \multicolumn{2}{c}{6.1} & \multicolumn{2}{c}{0.16} & \multicolumn{2}{c}{CIFAR10} & \multicolumn{2}{c}{Scalable} \\
    \multicolumn{2}{c}{RENAS~\cite{chen2019renas}} & \multicolumn{2}{c}{75.7} & \multicolumn{2}{c}{92.6} & \multicolumn{2}{c}{5.36} & \multicolumn{2}{c}{1.5} & \multicolumn{2}{c}{CIFAR10} & \multicolumn{2}{c}{Scalable} \\
    \multicolumn{2}{c}{CARS-C~\cite{yang2020cars}} & \multicolumn{2}{c}{75.2} & \multicolumn{2}{c}{92.5} & \multicolumn{2}{c}{5.1} & \multicolumn{2}{c}{0.4} & \multicolumn{2}{c}{CIFAR10} & \multicolumn{2}{c}{Scalable} \\
    \multicolumn{2}{c}{GDAS-NSAS-C~\cite{9247292}} & \multicolumn{2}{c}{74.1} & \multicolumn{2}{c}{–} & \multicolumn{2}{c}{5.2} & \multicolumn{2}{c}{0.4} & \multicolumn{2}{c}{CIFAR10} & \multicolumn{2}{c}{Scalable} \\
    \multicolumn{2}{c}{BayesNAS~\cite{zhou2019bayesnas}} & \multicolumn{2}{c}{73.5} & \multicolumn{2}{c}{91.1} & \multicolumn{2}{c}{3.9} & \multicolumn{2}{c}{0.2} & \multicolumn{2}{c}{CIFAR10} & \multicolumn{2}{c}{Scalable} \\
    \multicolumn{2}{c}{SETN~\cite{dong2019one}} & \multicolumn{2}{c}{74.3} & \multicolumn{2}{c}{92} & \multicolumn{2}{c}{5.4} & \multicolumn{2}{c}{1.8} & \multicolumn{2}{c}{CIFAR10} & \multicolumn{2}{c}{Scalable} \\
    \multicolumn{2}{c}{RandomNAS-NSAS-C~\cite{RandomNAS-NSAS}} & \multicolumn{2}{c}{74.5} & \multicolumn{2}{c}{–} & \multicolumn{2}{c}{5.4} & \multicolumn{2}{c}{0.7} & \multicolumn{2}{c}{CIFAR10} & \multicolumn{2}{c}{Scalable} \\
    \multicolumn{2}{c}{SNAS(mild)~\cite{xie2018snas}} & \multicolumn{2}{c}{72.7} & \multicolumn{2}{c}{90.8} & \multicolumn{2}{c}{4.3} & \multicolumn{2}{c}{1.5} & \multicolumn{2}{c}{CIFAR10} & \multicolumn{2}{c}{Scalable} \\
    \multicolumn{2}{c}{PDARTS~\cite{chen2019progressive}} & \multicolumn{2}{c}{75.3} & \multicolumn{2}{c}{92.5} & \multicolumn{2}{c}{5.1} & \multicolumn{2}{c}{0.3} & \multicolumn{2}{c}{CIFAR100} & \multicolumn{2}{c}{Scalable} \\
    \multicolumn{2}{c}{NSGANetV1-A3~\cite{lu2020multiobjective}} & \multicolumn{2}{c}{76.2} & \multicolumn{2}{c}{93.0} & \multicolumn{2}{c}{5} & \multicolumn{2}{c}{27} & \multicolumn{2}{c}{CIFAR100} & \multicolumn{2}{c}{Scalable} \\
    \multicolumn{2}{c}{TS-ENAS~\cite{zou2023ts}} & \multicolumn{2}{c}{75.6} & \multicolumn{2}{c}{-} & \multicolumn{2}{c}{3.6} & \multicolumn{2}{c}{0.9} & \multicolumn{2}{c}{CIFAR10} & \multicolumn{2}{c}{Hierarchical} \\
    \midrule
    \multicolumn{2}{c}{EPCNAS-C~\cite{9930866}} & \multicolumn{2}{c}{72.9†} & \multicolumn{2}{c}{91.5} & \multicolumn{2}{c}{3.02} & \multicolumn{2}{c}{1.17} & \multicolumn{2}{c}{ImageNet} & \multicolumn{2}{c}{Global} \\
    \multicolumn{2}{c}{FairNAS~\cite{9247292}} & \multicolumn{2}{c}{75.3†} & \multicolumn{2}{c}{–} & \multicolumn{2}{c}{4.6} & \multicolumn{2}{c}{12} & \multicolumn{2}{c}{ImageNet} & \multicolumn{2}{c}{Global} \\
    \multicolumn{2}{c}{PC-DARTS~\cite{xu2021partially}} & \multicolumn{2}{c}{75.8†} & \multicolumn{2}{c}{92.7} & \multicolumn{2}{c}{5.3} & \multicolumn{2}{c}{3.8} & \multicolumn{2}{c}{ImageNet} & \multicolumn{2}{c}{Scalable} \\
    \multicolumn{2}{c}{Proxyless NAS(GPU)~\cite{cai2018proxylessnas}} & \multicolumn{2}{c}{75.1†} & \multicolumn{2}{c}{92.5} & \multicolumn{2}{c}{7.1} & \multicolumn{2}{c}{8.3} & \multicolumn{2}{c}{ImageNet} & \multicolumn{2}{c}{Hierarchical} \\
    \multicolumn{2}{c}{MnasNet-A3~\cite{tan2019mnasnet}} & \multicolumn{2}{c}{76.7†} & \multicolumn{2}{c}{93.3} & \multicolumn{2}{c}{5.2} & \multicolumn{2}{c}{1600} & \multicolumn{2}{c}{ImageNet} & \multicolumn{2}{c}{Hierarchical} \\
    \multicolumn{2}{c}{MixNet-M~\cite{DBLP:journals/corr/abs-1907-09595}} & \multicolumn{2}{c}{\textbf{77†}} & \multicolumn{2}{c}{\textbf{93.3}} & \multicolumn{2}{c}{5} & \multicolumn{2}{c}{1600} & \multicolumn{2}{c}{ImageNet} & \multicolumn{2}{c}{Hierarchical} \\
    \midrule
    \multicolumn{2}{c}{MPAE-A} & \multicolumn{2}{c}{74.1} & \multicolumn{2}{c}{91.9} & \multicolumn{2}{c}{4.2} & \multicolumn{2}{c}{0.3} & \multicolumn{2}{c}{CIFAR10} & \multicolumn{2}{c}{Our} \\
    \multicolumn{2}{c}{MPAE-B} & \multicolumn{2}{c}{75.1} & \multicolumn{2}{c}{92.5} & \multicolumn{2}{c}{4.8} & \multicolumn{2}{c}{0.3} & \multicolumn{2}{c}{CIFAR10} & \multicolumn{2}{c}{Our} \\
    \multicolumn{2}{c}{MPAE-C} & \multicolumn{2}{c}{75.7} & \multicolumn{2}{c}{92.7} & \multicolumn{2}{c}{5.2} & \multicolumn{2}{c}{0.3} & \multicolumn{2}{c}{CIFAR10} & \multicolumn{2}{c}{Our} \\
    \bottomrule
    \end{tabular}%
  \caption{Overall comparison of MPAE with other NAS and OSNAS methods on the ILSVRC2012 dataset. The MPAE model is a architecture searched on the CIFAR-10 dataset. The symbol “†” in Table 2 indicates that the network was directly searched on ImageNet. The term “search dataset” refers to the dataset used for the neural architecture search and evaluation process.}
  \label{ImageNet}%
\end{table*}%

\subsection{Image Classification on CIFAR}
\paragraph{Search on CIFAR.} We split the CIFAR-10 and CIFAR-100 train set into two parts, i.e., 25,000 images for updating network parameters and 25,000 for verification architectures. The split strategy is the same as DARTS~\cite{liu2018darts}. We search for 500 epochs in total, and the parameter warmup stage lasts for the first $10\%$ epochs (50). After that, we initialize the population, which maintains 64 different architectures and gradually evolve them using proposed MPAE. We use MPAE to update architectures after the network parameters are updated for ten epochs.

\paragraph{Evaluate on CIFAR.}
After the MPAE search phase, we retrained the searched architectures on the CIFAR-10 and CIFAR-100 datasets with all the same training parameters as DARTS~\cite{liu2018darts}. As shown in Table I, on CIFAR-10 and CIFAR-100, MPAE outperforms most of its competitors in the Scalable category with similar parameters and a search cost of only 0.3 GDs. On the CIFAR-10 dataset, although NSGANetV1-A3 shows better classification accuracy than MPAE, the search cost of MPAE is much lower (only 0.3 GDs, much lower than 27 GDs). On CIFAR100, MPAE achieved a classification accuracy of $84.12\%$, surpassing its peer competitors in all Scalable categories considered in the experiment. Compared with peer competitors in the Global and Hierarchical categories, MPAE achieved competitive results on both CIFAR10 and CIFAR100 with lower search costs. Although the classification accuracy of Proxyless NAS is higher than MPAE-C on CIFAR10, MPAE-C has fewer parameters (3.7M $<$ 5.7M) and lower search cost (only 0.3 GDs, much lower than 1500 GDs). On CIFAR100, MPAE surpassed all Global and Hierarchical competitors considered in the experiment in both accuracy ($\%$) and search time (GDs).

\subsection{Image Classification on ImageNet}
\paragraph{Training Details.} We evaluate the transferability of the architectures obtained from the CIFAR-10 dataset by training them on the ImageNet dataset. We use Nvidia A100 to train the models, and the batch size is 256. We train 250 epochs in total. The learning rate is 0.1 with a linear decay scheduler. Momentum is 0.9, and weight decay is 3e-5. Label smooth is also used with a smooth ratio of 0.1.

\paragraph{Evaluate on ImageNet.} Table 2 summarizes the results of our model and other automated search networks on the ImageNet validation set. We categorize other search algorithms into two groups based on whether the search dataset is ImageNet or not. “Non-ImageNet” search dataset refers to methods that search for networks on small datasets and then transfer them to larger datasets. In comparison to other non-ImageNet datasets, the MPAE model exhibits significant advantages. Only the NSGANetV1-A3~\cite{lu2020multiobjective} model has a slightly higher accuracy ($0.5\%$) than MPAE, but its search cost is 90 times more than MPAE. Methods that require searching on the ImageNet dataset demand even more search time or computational resources. For example, although MixNet-M~\cite{DBLP:journals/corr/abs-1907-09595} and MnasNet-A3~\cite{tan2019mnasnet} models have higher accuracy than MPAE, they consume far more GPU-days than our approach (1600 GDs $>>$ 0.3 GDs). While other methods using ImageNet as the search dataset have a search cost of around 10 GDs, their performance is similar to or lower than MPAE. Therefore, MPAE remains effective in this regard, and it ensures module diversity with a smaller search cost, thereby guaranteeing model performance.

\begin{figure*}
  \centering
  \begin{subfigure}{0.495\textwidth}
    \includegraphics[width=\textwidth]{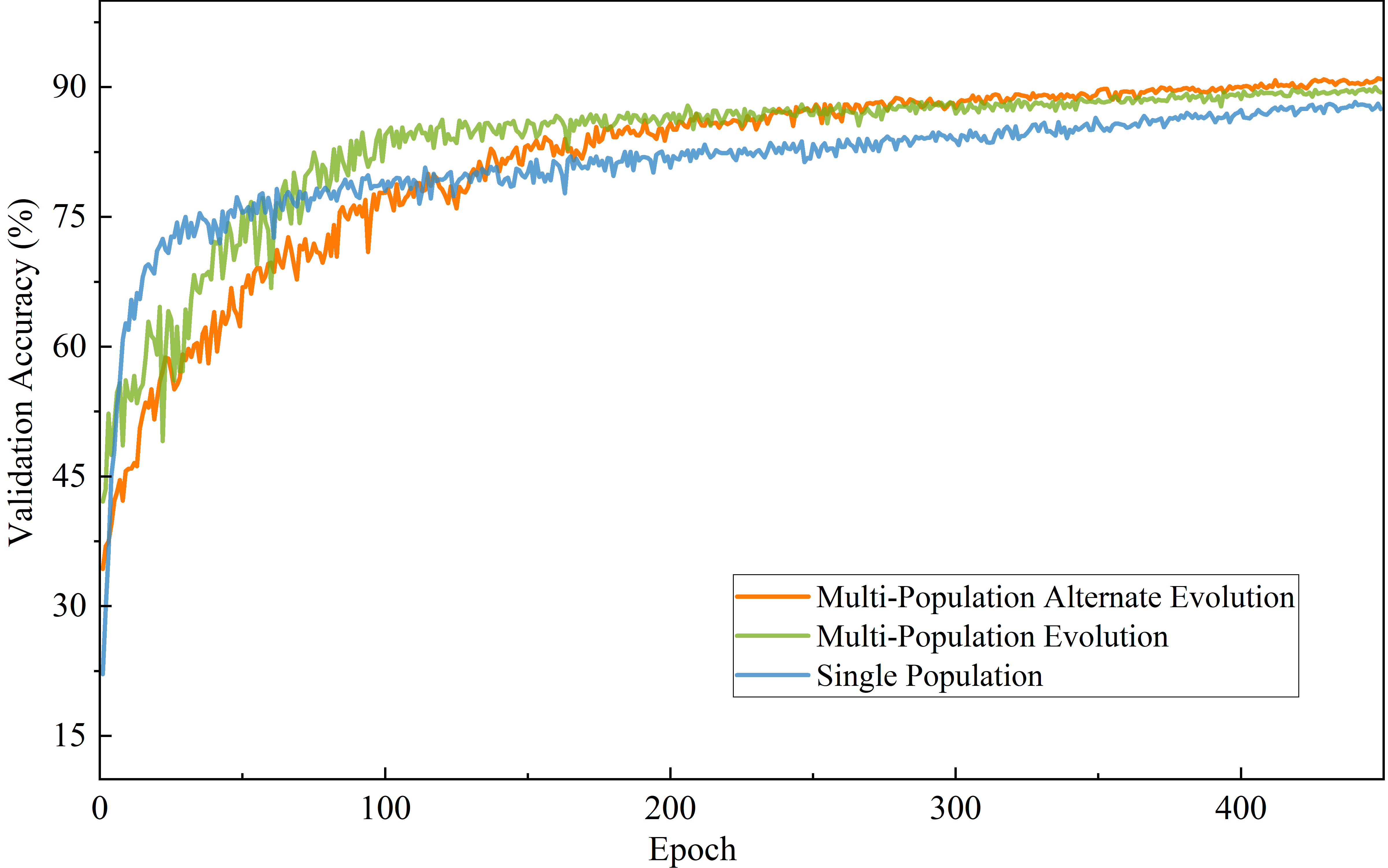}
    \caption{}
    \label{fig:short-a}
  \end{subfigure}
  \hfill
  \begin{subfigure}{0.495\textwidth}
    \includegraphics[width=\textwidth]{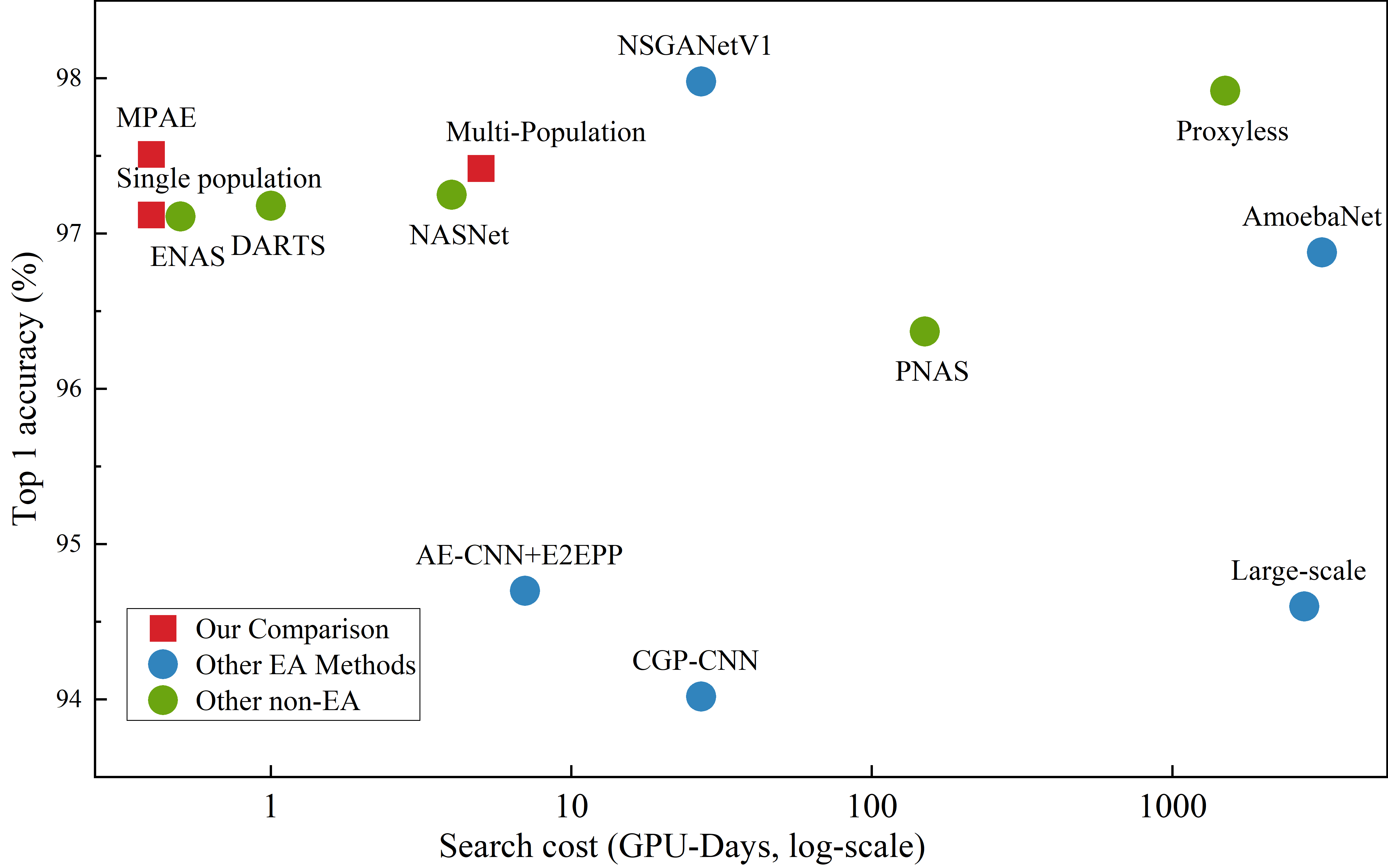}
    \caption{}
    \label{fig:short-b}
  \end{subfigure}
  \caption{Search efficiency comparison between MPAE and other baselines in terms of (a) validation accuracy and (b) required compute time in GPU-Days. The search cost is measured on CIFAR-10 for most methods, except Block-QNN, where the CIFAR-100 dataset is used for.}
  \label{fig:short}
\end{figure*}

\section{Ablation Study}
\begin{figure}[tb]
\centering
\subfloat[100 Epoch]{\includegraphics[width=0.505\linewidth]
{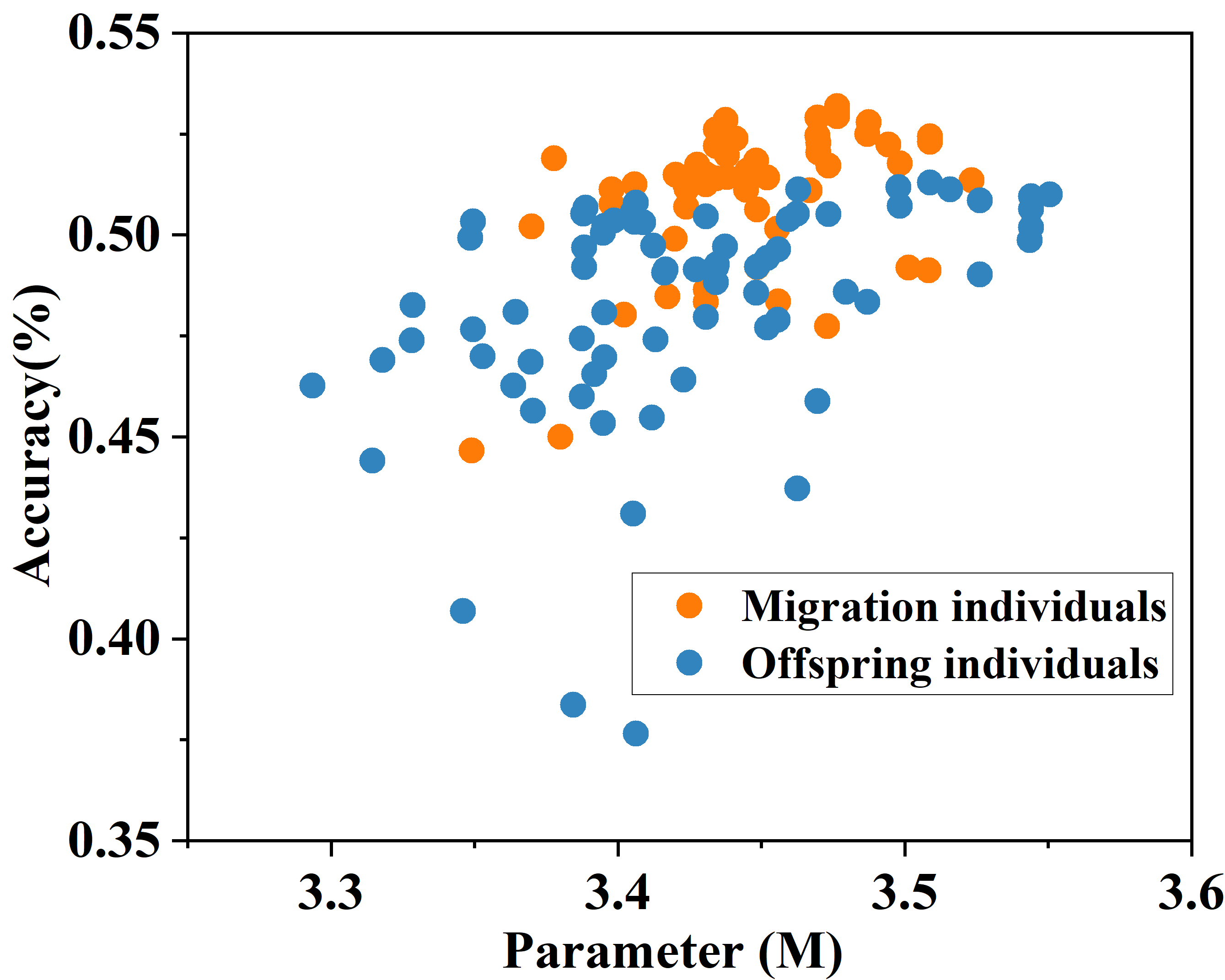}}
\subfloat[200 Epoch]{\includegraphics[width=0.5\linewidth]{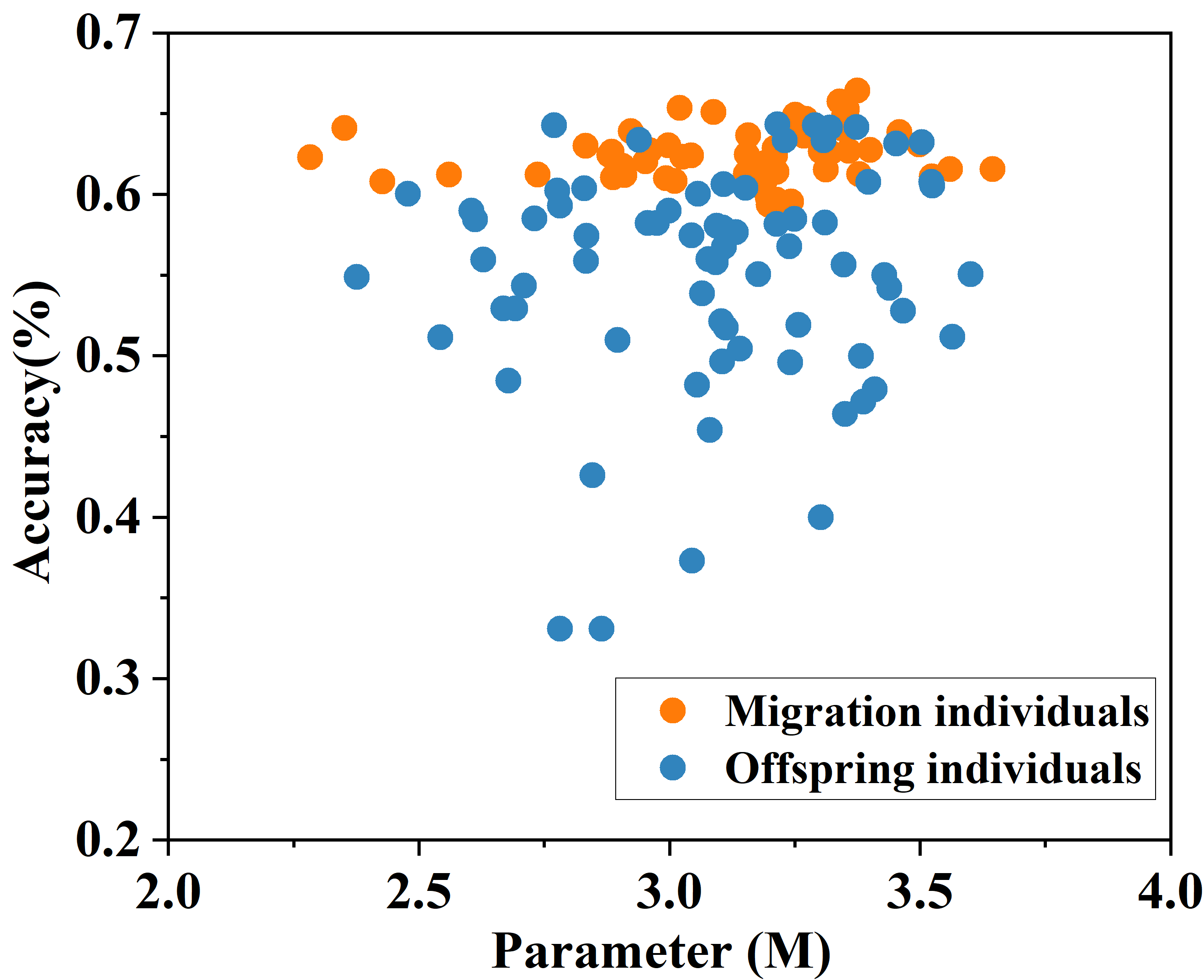}}
\\
\subfloat[300 Epoch]{\includegraphics[width=0.501\linewidth]{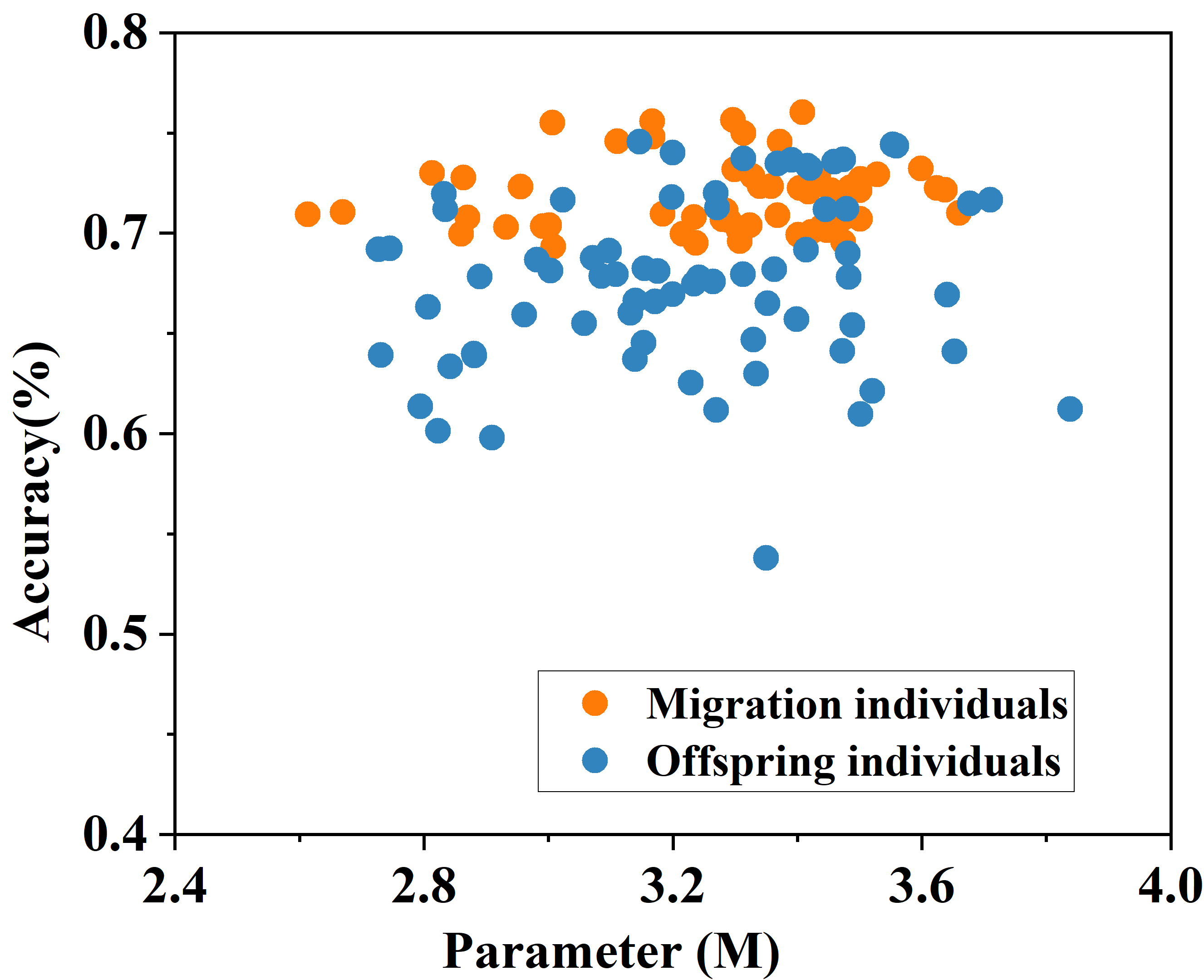}}
\subfloat[400 Epoch]{\includegraphics[width=0.5\linewidth]{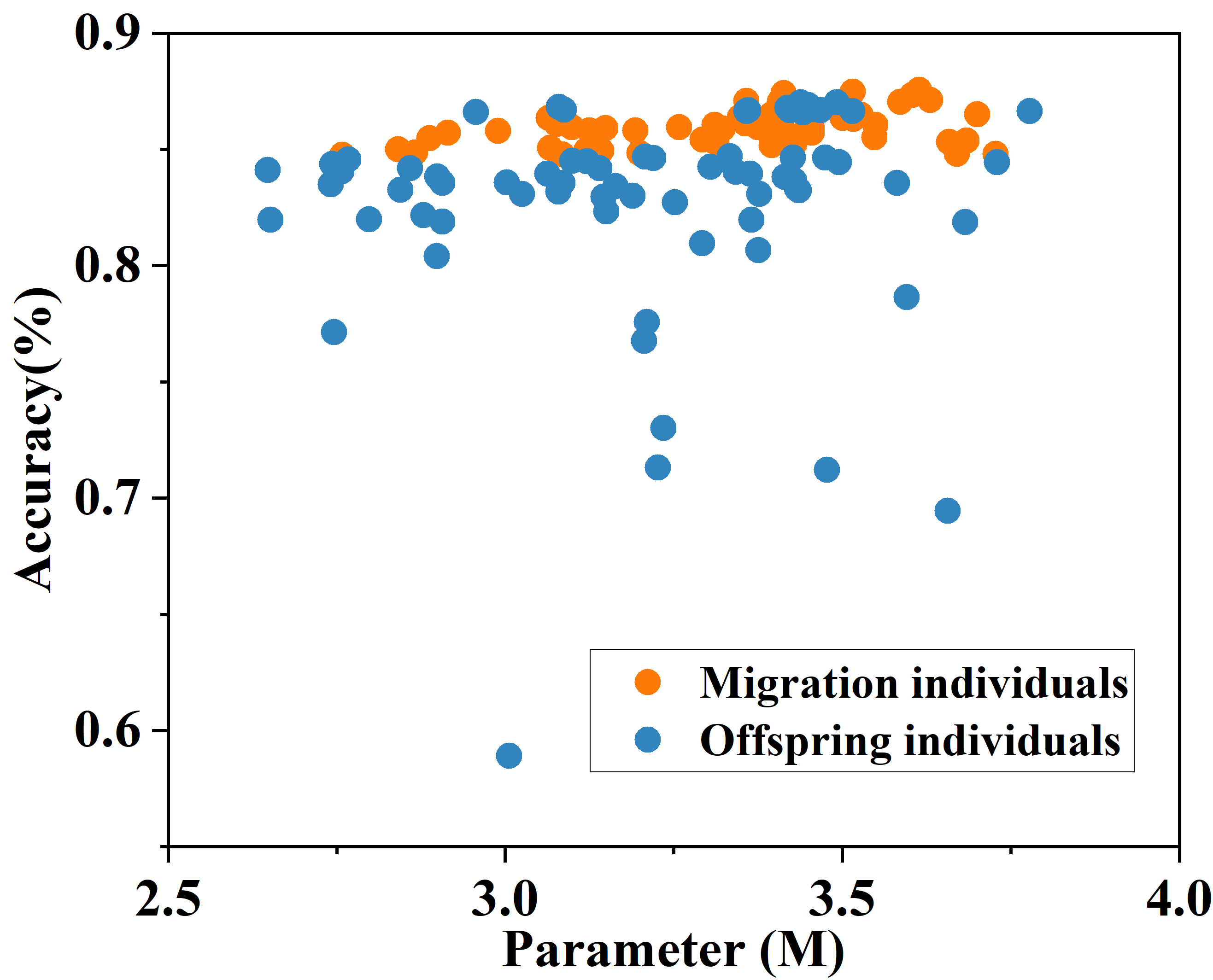}}
\caption{Comparison graph of migrated individuals and offspring individuals at various stages of the MPAE evolutionary process.}
\label{fig_6}
\end{figure}
\paragraph{Effectiveness of Multi-Populations Alternate Evolution:} Comparing the contributions of different NAS algorithms in terms of the search stage can be challenging and ambiguous due to the significant differences in the search space and training procedures used during the search process. Therefore, we use multi-population coevolution and single-population global search as comparisons with our MPAE algorithm to demonstrate the efficiency of the search stage in MPAE. All three methods utilize the same search space and performance estimation strategy. The results shown in Figure 3(a) indicate that MPAE is capable of finding more accurate architectures compared to multi-population coevolution and single-population global search.
In addition to accuracy metrics, another important aspect of demonstrating NAS efficiency is the computational complexity of the methods. Since performing a theoretical analysis of the computational complexity for different NAS methods is challenging, we compare the computational time (GPU-days) spent by each method on the GPU to derive the reported architectures. Running MPAE once on the CIFAR-10 dataset takes approximately 0.3 GPU-days to complete (average of five runs). From the search cost comparison in Figure 3(b), we observe that our proposed algorithm is faster in identifying a set of architectures compared to many other methods and the obtained architecture set exhibits higher performance.

\paragraph{Effectiveness of Migration Mechanism:}
To demonstrate the effectiveness of the proposed population migration mechanism, we compare individuals sampled uniformly from four stages (100, 200, 300, 400 epochs) of the population evolution process. Figure 4 displays the results of comparing individuals migrated from other populations with individuals generated through population crossover and mutation at each stage. It can be observed that the migrated individuals exhibit higher generalization accuracy, while the offspring individuals have a wider search range. MPAE improves population convergence through migrated individuals, while it enhances population diversity through offspring individuals.

\section{Conclusion}
This paper proposed the multiple population alternate evolution framework that mitigated the conflict between search costs and module diversity by achieving module diversification with lower search costs. We first simplify the search space into multiple search subspaces and then use multiple populations to alternately search within these subspaces. This significantly reduces the complexity of the problem and improves search efficiency. We also introduce a population migration mechanism, which aims to accelerate the evolution process by utilizing the knowledge and experience retained by each population. Furthermore, our experiments demonstrate that the performance of migrated individuals generally surpasses that of offspring generated by the populations. Our MPAE algorithm achieves state-of-the-art results with only 0.3 GPU days of search, it demonstrates outstanding performance in both search speed and search efficiency.

\section*{Acknowledgments}
This work was supported in part by the National Natural Science Foundation of China (Grant No. 62276224, Grant No. 6230073173), in part by the Natural Science Foundation of Hunan Province, China (Grant No. 2022JJ40452)

\bibliographystyle{named}
\bibliography{ijcai24}

\begin{thebibliography}{}

\bibitem[\protect\citeauthoryear{Bender \bgroup \em et al.\egroup }{2018}]{bender2018understanding}
Gabriel Bender, Pieter-Jan Kindermans, Barret Zoph, Vijay Vasudevan, and Quoc Le.
\newblock Understanding and simplifying one-shot architecture search.
\newblock In {\em International conference on machine learning}, pages 550--559. PMLR, 2018.

\bibitem[\protect\citeauthoryear{Benyahia \bgroup \em et al.\egroup }{2019}]{pmlr-v97-benyahia19a}
Yassine Benyahia, Kaicheng Yu, Kamil~Bennani Smires, Martin Jaggi, Anthony~C. Davison, Mathieu Salzmann, and Claudiu Musat.
\newblock Overcoming multi-model forgetting.
\newblock In Kamalika Chaudhuri and Ruslan Salakhutdinov, editors, {\em Proceedings of the 36th International Conference on Machine Learning}, volume~97 of {\em Proceedings of Machine Learning Research}, pages 594--603. PMLR, 09--15 Jun 2019.

\bibitem[\protect\citeauthoryear{Cai \bgroup \em et al.\egroup }{2019}]{cai2018proxylessnas}
Han Cai, Ligeng Zhu, and Song Han.
\newblock Proxyless{NAS}: Direct neural architecture search on target task and hardware.
\newblock In {\em International Conference on Learning Representations}, 2019.

\bibitem[\protect\citeauthoryear{Chen \bgroup \em et al.\egroup }{2019a}]{chen2019progressive}
Xin Chen, Lingxi Xie, Jun Wu, and Qi~Tian.
\newblock Progressive differentiable architecture search: Bridging the depth gap between search and evaluation.
\newblock In {\em Proceedings of the IEEE/CVF international conference on computer vision}, pages 1294--1303, 2019.

\bibitem[\protect\citeauthoryear{Chen \bgroup \em et al.\egroup }{2019b}]{chen2019renas}
Yukang Chen, Gaofeng Meng, Qian Zhang, Shiming Xiang, Chang Huang, Lisen Mu, and Xinggang Wang.
\newblock Renas: Reinforced evolutionary neural architecture search.
\newblock In {\em Proceedings of the IEEE/CVF conference on computer vision and pattern recognition}, pages 4787--4796, 2019.

\bibitem[\protect\citeauthoryear{Dong and Yang}{2019}]{dong2019one}
Xuanyi Dong and Yi~Yang.
\newblock One-shot neural architecture search via self-evaluated template network.
\newblock In {\em Proceedings of the IEEE/CVF International Conference on Computer Vision}, pages 3681--3690, 2019.

\bibitem[\protect\citeauthoryear{Girish \bgroup \em et al.\egroup }{2019}]{girish2019unsupervised}
Deeptha Girish, Vineeta Singh, and Anca~L Ralescu.
\newblock Unsupervised clustering based understanding of cnn.
\newblock In {\em CVPR Workshops}, pages 9--11, 2019.

\bibitem[\protect\citeauthoryear{He \bgroup \em et al.\egroup }{2016}]{he2016deep}
Kaiming He, Xiangyu Zhang, Shaoqing Ren, and Jian Sun.
\newblock Deep residual learning for image recognition.
\newblock In {\em Proceedings of the IEEE conference on computer vision and pattern recognition}, pages 770--778, 2016.

\bibitem[\protect\citeauthoryear{Huang \bgroup \em et al.\egroup }{2023}]{9930866}
Junhao Huang, Bing Xue, Yanan Sun, Mengjie Zhang, and Gary~G. Yen.
\newblock Particle swarm optimization for compact neural architecture search for image classification.
\newblock {\em IEEE Transactions on Evolutionary Computation}, 27(5):1298--1312, 2023.

\bibitem[\protect\citeauthoryear{Klyuchnikov \bgroup \em et al.\egroup }{2022}]{klyuchnikov2022bench}
Nikita Klyuchnikov, Ilya Trofimov, Ekaterina Artemova, Mikhail Salnikov, Maxim Fedorov, Alexander Filippov, and Evgeny Burnaev.
\newblock Nas-bench-nlp: neural architecture search benchmark for natural language processing.
\newblock {\em IEEE Access}, 10:45736--45747, 2022.

\bibitem[\protect\citeauthoryear{Liang \bgroup \em et al.\egroup }{2021}]{liang2021opanas}
Tingting Liang, Yongtao Wang, Zhi Tang, Guosheng Hu, and Haibin Ling.
\newblock Opanas: One-shot path aggregation network architecture search for object detection.
\newblock In {\em Proceedings of the IEEE/CVF conference on computer vision and pattern recognition}, pages 10195--10203, 2021.

\bibitem[\protect\citeauthoryear{Liu \bgroup \em et al.\egroup }{2018a}]{liu2018progressive}
Chenxi Liu, Barret Zoph, Maxim Neumann, Jonathon Shlens, Wei Hua, Li-Jia Li, Li~Fei-Fei, Alan Yuille, Jonathan Huang, and Kevin Murphy.
\newblock Progressive neural architecture search.
\newblock In {\em Proceedings of the European conference on computer vision (ECCV)}, pages 19--34, 2018.

\bibitem[\protect\citeauthoryear{Liu \bgroup \em et al.\egroup }{2018b}]{liu2018darts}
Hanxiao Liu, Karen Simonyan, and Yiming Yang.
\newblock Darts: Differentiable architecture search.
\newblock In {\em International Conference on Learning Representations}, 2018.

\bibitem[\protect\citeauthoryear{Lu \bgroup \em et al.\egroup }{2020}]{lu2020multiobjective}
Zhichao Lu, Ian Whalen, Yashesh Dhebar, Kalyanmoy Deb, Erik~D Goodman, Wolfgang Banzhaf, and Vishnu~Naresh Boddeti.
\newblock Multiobjective evolutionary design of deep convolutional neural networks for image classification.
\newblock {\em IEEE Transactions on Evolutionary Computation}, 25(2):277--291, 2020.

\bibitem[\protect\citeauthoryear{Pham \bgroup \em et al.\egroup }{2018}]{pham2018efficient}
Hieu Pham, Melody Guan, Barret Zoph, Quoc Le, and Jeff Dean.
\newblock Efficient neural architecture search via parameters sharing.
\newblock In {\em International conference on machine learning}, pages 4095--4104. PMLR, 2018.

\bibitem[\protect\citeauthoryear{Real \bgroup \em et al.\egroup }{2017}]{real2017large}
Esteban Real, Sherry Moore, Andrew Selle, Saurabh Saxena, Yutaka~Leon Suematsu, Jie Tan, Quoc~V Le, and Alexey Kurakin.
\newblock Large-scale evolution of image classifiers.
\newblock In {\em International Conference on Machine Learning}, pages 2902--2911. PMLR, 2017.

\bibitem[\protect\citeauthoryear{Real \bgroup \em et al.\egroup }{2019}]{real2019regularized}
Esteban Real, Alok Aggarwal, Yanping Huang, and Quoc~V Le.
\newblock Regularized evolution for image classifier architecture search.
\newblock In {\em Proceedings of the aaai conference on artificial intelligence}, volume~33, pages 4780--4789, 2019.

\bibitem[\protect\citeauthoryear{Sun \bgroup \em et al.\egroup }{2019a}]{sun2019surrogate}
Yanan Sun, Handing Wang, Bing Xue, Yaochu Jin, Gary~G Yen, and Mengjie Zhang.
\newblock Surrogate-assisted evolutionary deep learning using an end-to-end random forest-based performance predictor.
\newblock {\em IEEE Transactions on Evolutionary Computation}, 24(2):350--364, 2019.

\bibitem[\protect\citeauthoryear{Sun \bgroup \em et al.\egroup }{2019b}]{sun2019completely}
Yanan Sun, Bing Xue, Mengjie Zhang, and Gary~G Yen.
\newblock Completely automated cnn architecture design based on blocks.
\newblock {\em IEEE transactions on neural networks and learning systems}, 31(4):1242--1254, 2019.

\bibitem[\protect\citeauthoryear{Sun \bgroup \em et al.\egroup }{2019c}]{sun2019evolving}
Yanan Sun, Bing Xue, Mengjie Zhang, and Gary~G Yen.
\newblock Evolving deep convolutional neural networks for image classification.
\newblock {\em IEEE Transactions on Evolutionary Computation}, 24(2):394--407, 2019.

\bibitem[\protect\citeauthoryear{Sun \bgroup \em et al.\egroup }{2020}]{sun2020automatically}
Yanan Sun, Bing Xue, Mengjie Zhang, Gary~G Yen, and Jiancheng Lv.
\newblock Automatically designing cnn architectures using the genetic algorithm for image classification.
\newblock {\em IEEE transactions on cybernetics}, 50(9):3840--3854, 2020.

\bibitem[\protect\citeauthoryear{Szegedy \bgroup \em et al.\egroup }{2017}]{szegedy2017inception}
Christian Szegedy, Sergey Ioffe, Vincent Vanhoucke, and Alexander Alemi.
\newblock Inception-v4, inception-resnet and the impact of residual connections on learning.
\newblock In {\em Proceedings of the AAAI conference on artificial intelligence}, volume~31, 2017.

\bibitem[\protect\citeauthoryear{Tan and Le}{2019}]{DBLP:journals/corr/abs-1907-09595}
Mingxing Tan and Quoc~V. Le.
\newblock Mixconv: Mixed depthwise convolutional kernels.
\newblock {\em CoRR}, abs/1907.09595, 2019.

\bibitem[\protect\citeauthoryear{Tan \bgroup \em et al.\egroup }{2019}]{tan2019mnasnet}
Mingxing Tan, Bo~Chen, Ruoming Pang, Vijay Vasudevan, Mark Sandler, Andrew Howard, and Quoc~V Le.
\newblock Mnasnet: Platform-aware neural architecture search for mobile.
\newblock In {\em Proceedings of the IEEE/CVF conference on computer vision and pattern recognition}, pages 2820--2828, 2019.

\bibitem[\protect\citeauthoryear{Tian \bgroup \em et al.\egroup }{2020}]{tian2020multipopulation}
Ye~Tian, Ruchen Liu, Xingyi Zhang, Haiping Ma, Kay~Chen Tan, and Yaochu Jin.
\newblock A multipopulation evolutionary algorithm for solving large-scale multimodal multiobjective optimization problems.
\newblock {\em IEEE Transactions on Evolutionary Computation}, 25(3):405--418, 2020.

\bibitem[\protect\citeauthoryear{Wu \bgroup \em et al.\egroup }{2019}]{wu2019fbnet}
Bichen Wu, Xiaoliang Dai, Peizhao Zhang, Yanghan Wang, Fei Sun, Yiming Wu, Yuandong Tian, Peter Vajda, Yangqing Jia, and Kurt Keutzer.
\newblock Fbnet: Hardware-aware efficient convnet design via differentiable neural architecture search.
\newblock In {\em Proceedings of the IEEE/CVF Conference on Computer Vision and Pattern Recognition}, pages 10734--10742, 2019.

\bibitem[\protect\citeauthoryear{Xie and Yuille}{2017}]{xie2017genetic}
Lingxi Xie and Alan Yuille.
\newblock Genetic cnn.
\newblock In {\em Proceedings of the IEEE international conference on computer vision}, pages 1379--1388, 2017.

\bibitem[\protect\citeauthoryear{Xie \bgroup \em et al.\egroup }{2019}]{xie2018snas}
Sirui Xie, Hehui Zheng, Chunxiao Liu, and Liang Lin.
\newblock {SNAS}: stochastic neural architecture search.
\newblock In {\em International Conference on Learning Representations}, 2019.

\bibitem[\protect\citeauthoryear{Xu \bgroup \em et al.\egroup }{2021}]{xu2021partially}
Yuhui Xu, Lingxi Xie, Wenrui Dai, Xiaopeng Zhang, Xin Chen, Guo-Jun Qi, Hongkai Xiong, and Qi~Tian.
\newblock Partially-connected neural architecture search for reduced computational redundancy.
\newblock {\em IEEE Transactions on Pattern Analysis and Machine Intelligence}, 43(9):2953--2970, 2021.

\bibitem[\protect\citeauthoryear{Yang \bgroup \em et al.\egroup }{2020}]{yang2020cars}
Zhaohui Yang, Yunhe Wang, Xinghao Chen, Boxin Shi, Chao Xu, Chunjing Xu, Qi~Tian, and Chang Xu.
\newblock Cars: Continuous evolution for efficient neural architecture search.
\newblock In {\em Proceedings of the IEEE/CVF Conference on Computer Vision and Pattern Recognition}, pages 1829--1838, 2020.

\bibitem[\protect\citeauthoryear{Yu \bgroup \em et al.\egroup }{2019}]{yu2019evaluating}
Kaicheng Yu, Christian Sciuto, Martin Jaggi, Claudiu Musat, and Mathieu Salzmann.
\newblock Evaluating the search phase of neural architecture search.
\newblock In {\em International Conference on Learning Representations}, 2019.

\bibitem[\protect\citeauthoryear{Zeiler and Fergus}{2014}]{zeiler2014visualizing}
Matthew~D Zeiler and Rob Fergus.
\newblock Visualizing and understanding convolutional networks.
\newblock In {\em Computer Vision--ECCV 2014: 13th European Conference, Zurich, Switzerland, September 6-12, 2014, Proceedings, Part I 13}, pages 818--833. Springer, 2014.

\bibitem[\protect\citeauthoryear{Zhang \bgroup \em et al.\egroup }{2020}]{zhang2020you}
Xinbang Zhang, Zehao Huang, Naiyan Wang, Shiming Xiang, and Chunhong Pan.
\newblock You only search once: Single shot neural architecture search via direct sparse optimization.
\newblock {\em IEEE Transactions on Pattern Analysis and Machine Intelligence}, 43(9):2891--2904, 2020.

\bibitem[\protect\citeauthoryear{Zhang \bgroup \em et al.\egroup }{2021a}]{9268174}
Haoyu Zhang, Yaochu Jin, Ran Cheng, and Kuangrong Hao.
\newblock Efficient evolutionary search of attention convolutional networks via sampled training and node inheritance.
\newblock {\em IEEE Transactions on Evolutionary Computation}, 25(2):371--385, 2021.

\bibitem[\protect\citeauthoryear{Zhang \bgroup \em et al.\egroup }{2021b}]{9247292}
Miao Zhang, Huiqi Li, Shirui Pan, Xiaojun Chang, Chuan Zhou, Zongyuan Ge, and Steven Su.
\newblock One-shot neural architecture search: Maximising diversity to overcome catastrophic forgetting.
\newblock {\em IEEE Transactions on Pattern Analysis and Machine Intelligence}, 43(9):2921--2935, 2021.

\bibitem[\protect\citeauthoryear{Zhang \bgroup \em et al.\egroup }{2021c}]{RandomNAS-NSAS}
Miao Zhang, Huiqi Li, Shirui Pan, Xiaojun Chang, Chuan Zhou, Zongyuan Ge, and Steven Su.
\newblock One-shot neural architecture search: Maximising diversity to overcome catastrophic forgetting.
\newblock {\em IEEE Transactions on Pattern Analysis and Machine Intelligence}, 43(9):2921--2935, 2021.

\bibitem[\protect\citeauthoryear{Zhang \bgroup \em et al.\egroup }{2022}]{zhang2022evolutionary}
Haoyu Zhang, Yaochu Jin, and Kuangrong Hao.
\newblock Evolutionary search for complete neural network architectures with partial weight sharing.
\newblock {\em IEEE transactions on evolutionary computation}, 26(5):1072--1086, 2022.

\bibitem[\protect\citeauthoryear{Zheng \bgroup \em et al.\egroup }{2019}]{zheng2019multinomial}
Xiawu Zheng, Rongrong Ji, Lang Tang, Baochang Zhang, Jianzhuang Liu, and Qi~Tian.
\newblock Multinomial distribution learning for effective neural architecture search.
\newblock In {\em Proceedings of the IEEE/CVF International Conference on Computer Vision}, pages 1304--1313, 2019.

\bibitem[\protect\citeauthoryear{Zhong \bgroup \em et al.\egroup }{2020}]{zhong2020blockqnn}
Zhao Zhong, Zichen Yang, Boyang Deng, Junjie Yan, Wei Wu, Jing Shao, and Cheng-Lin Liu.
\newblock Blockqnn: Efficient block-wise neural network architecture generation.
\newblock {\em IEEE transactions on pattern analysis and machine intelligence}, 43(7):2314--2328, 2020.

\bibitem[\protect\citeauthoryear{Zhou \bgroup \em et al.\egroup }{2019}]{zhou2019bayesnas}
Hongpeng Zhou, Minghao Yang, Jun Wang, and Wei Pan.
\newblock Bayesnas: A bayesian approach for neural architecture search.
\newblock In {\em International conference on machine learning}, pages 7603--7613. PMLR, 2019.

\bibitem[\protect\citeauthoryear{Zoph and Le}{2016}]{zoph2016neural}
Barret Zoph and Quoc Le.
\newblock Neural architecture search with reinforcement learning.
\newblock In {\em International Conference on Learning Representations}, 2016.

\bibitem[\protect\citeauthoryear{Zoph \bgroup \em et al.\egroup }{2018}]{zoph2018learning}
Barret Zoph, Vijay Vasudevan, Jonathon Shlens, and Quoc~V Le.
\newblock Learning transferable architectures for scalable image recognition.
\newblock In {\em Proceedings of the IEEE conference on computer vision and pattern recognition}, pages 8697--8710, 2018.

\bibitem[\protect\citeauthoryear{Zou \bgroup \em et al.\egroup }{2023}]{zou2023ts}
Juan Zou, Shenghong Wu, Yizhang Xia, Weiwei Jiang, Zeping Wu, and Jinhua Zheng.
\newblock Ts-enas: Two-stage evolution for cell-based network architecture search.
\newblock {\em arXiv preprint arXiv:2310.09525}, 2023.

\end{thebibliography}

\end{document}